\documentclass[letterpaper, 10 pt, journal, twoside]{IEEEtran}  

\IEEEoverridecommandlockouts                              

\usepackage{graphicx}
\usepackage{amsmath, amssymb}
\usepackage{makecell}
\usepackage{multirow}
\usepackage{subfigure}
\usepackage{url}
\usepackage{threeparttable}
\usepackage{cite}

\title{\LARGE \bf
EfficientGrasp: A Unified Data-Efficient Learning to\\ Grasp Method for Multi-fingered Robot Hands 
}
\author{Kelin~Li, Nicholas~Baron, Xian~Zhang, and~Nicolas~Rojas,~\IEEEmembership{Member,~IEEE}

\thanks{Kelin Li, Xian Zhang and Nicolas Rojas are with the REDS Lab, Dyson School of Design Engineering, Imperial College London, 25 Exhibition Road, London, SW7 2DB, UK
{\tt\footnotesize (k.li20, xian.zhang17, n.rojas)@imperial.ac.uk}. Nicholas Baron was with the REDS Lab. He is currently with Ocado Technology, Trident Place, Hatfield, AL10 9UL, UK
{\tt\footnotesize nicholas.baron@ocado.com}}

}

\begin{document}

\maketitle

\begin{abstract}

Autonomous grasping of novel objects that are previously unseen to a robot is an ongoing challenge in robotic manipulation. In the last decades, many approaches have been presented to address this problem for specific robot hands. The UniGrasp framework, introduced recently, has the ability to generalize to different types of robotic grippers; however, this method does not work on grippers with closed-loop constraints and is data-inefficient when applied to robot hands with multi-grasp configurations. In this paper, we present \textit{EfficientGrasp}, a generalized grasp synthesis and gripper control method that is independent of gripper model specifications. \textit{EfficientGrasp} utilizes a gripper workspace feature rather than UniGrasp's gripper attribute inputs. This reduces memory use by 81.7\% during training and makes it possible to generalize to more types of grippers, such as grippers with closed-loop constraints. The effectiveness of \textit{EfficientGrasp} is evaluated by conducting object grasping experiments both in simulation and real-world; results show that the proposed method also outperforms UniGrasp when considering only grippers without closed-loop constraints. In these cases, \textit{EfficientGrasp} shows 9.85\% higher accuracy in generating contact points and 3.10\% higher grasping success rate in simulation. The real-world experiments are conducted with a gripper with closed-loop constraints, which UniGrasp fails to handle while \textit{EfficientGrasp} achieves a success rate of 83.3\%. The main causes of grasping failures of the proposed method are analyzed, highlighting ways of enhancing grasp performance. 

\end{abstract}

\section{Introduction}
\label{section1}

\IEEEPARstart{T}{he} ability to grasp a large variety of objects is one of the most important capabilities for robot hands to improve in order to be used more in our daily life. Recent works have proposed multiple algorithms to address this, particularly for cases in which the objects are unknown or have not been seen by the robot previously. Among recent approaches for grasp synthesis of novel objects, visual data-driven methods are some of the most popular \cite{joshi2020robotic,ni2020pointnet++,liang2019pointnetgpd,boroushaki2021robotic,bousmalis2018using,hang2016hierarchical,bohg2013data}. These techniques are characterised in general by the use of large amounts of pre-trained visual data to be able to generalize to novel, unseen objects. However, these methods are almost exclusively trained for a specific robot gripper type and are not usually directly applicable to new types of grippers with different geometries or configurations.

\begin{figure*}[t]
    \centering
    \includegraphics[width=0.95\textwidth,height=0.394\textwidth]{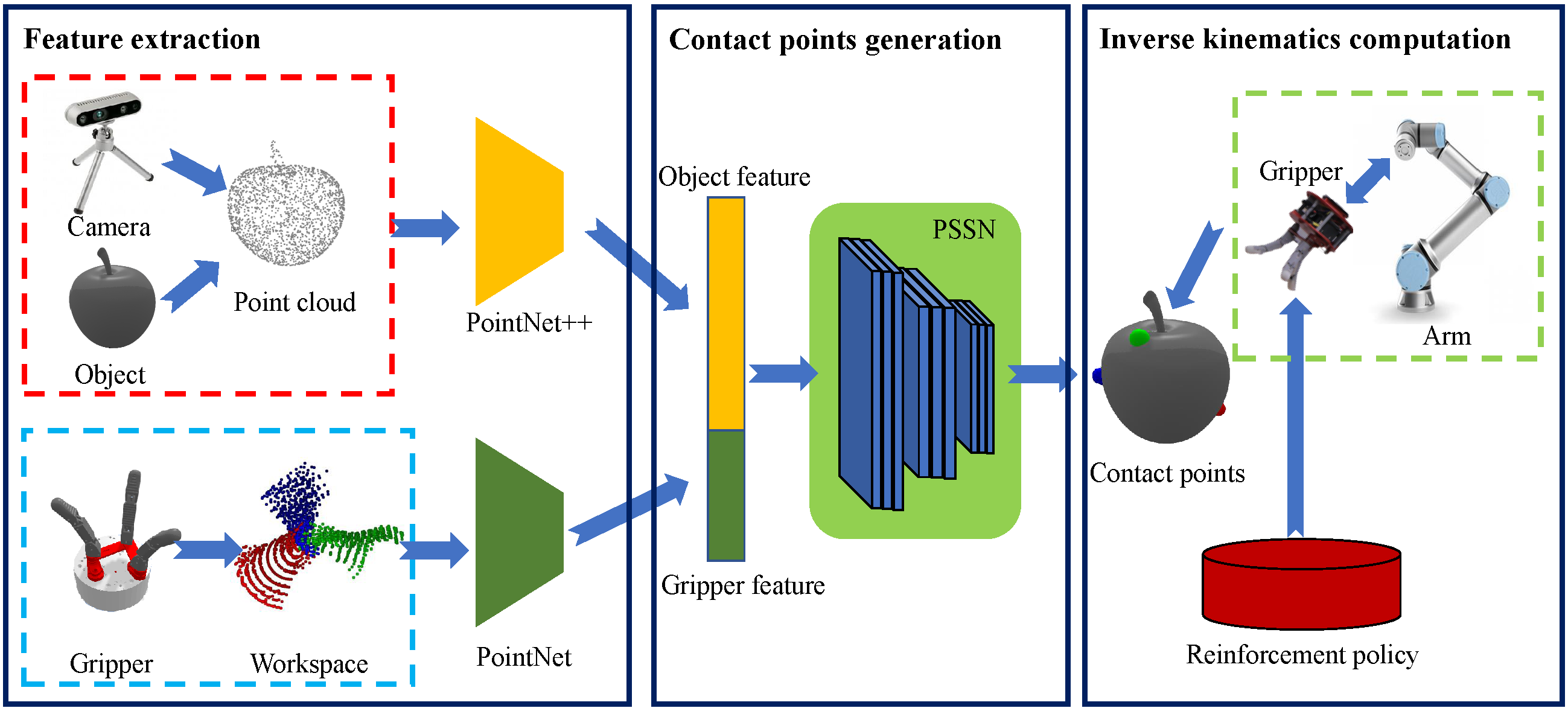}
    \caption{Flowchart of the proposed EfficientGrasp. It consists of three phases, namely feature extraction, contact points generation, and inverse kinematics computation. In the feature extraction phase, the object feature and gripper workspace feature are extracted by PointNet. Then the two features are concatenated and fed into a trained PSSN to generate contact points on the object. Finally, the gripper IK is computed by reinforcement policy, and the robot arm is controlled to approach the contact points.}
    \label{fig:flowchart}
\end{figure*}

 A notable exception to the lack of grasping methods which generalize to new grippers is the UniGrasp framework~\cite{shao2020UniGrasp}. This method takes a point cloud of a novel object and gripper, from a model of the gripper specified in Unified Robot Description Format (URDF), as inputs for obtaining contact points that satisfy force closure conditions. The inverse kinematics of the gripper is then solved using the Rigid Body Dynamics Library (RBDL)~\cite{felis2017rbdl}, which is a library that can calculate the robot inverse kinematics based on the input URDF model of the gripper. UniGrasp has been proven to be generalizable to novel objects and gripper types, provided that the gripper can be represented by a URDF (i.e. does not have closed-loop kinematic chains). However, the method becomes data-inefficient and cumbersome when applied to grippers in which `most open' and `most closed' gripper operation modes are ambiguous, as in the Robotiq-3F gripper, the BarrettHand, or the RUTH hand~\cite{luruth}. Moreover, UniGrasp cannot work on grippers with closed-loop constraints (e.g., the RUTH hand, the integrated linkage-driven hand~\cite{kim2021integrated}) since a URDF cannot represent mechanisms with this characteristic.  
 
 In this paper, we present an improved data-driven grasp synthesis and gripper control approach named \textit{EfficientGrasp}, which can address the data-inefficient issues mentioned above. This method generalizes efficiently to novel multi-finger grippers regardless of geometry and kinematics, and does not use hand model specifications. EfficientGrasp uses fingertip workspace information as the gripper attribute input, instead of a URDF model for instance, and uses an object point cloud as training input. A deep neural network, a point set  selection  network  (PSSN)~\cite{qi2017pointnet++}, is used to generate force-closure contact points on given novel objects from a concatenation of object-gripper features. A model-free reinforcement learning method, a soft actor-critic~\cite{haarnoja2018soft}, is then used for solving the inverse kinematics of the gripper. A flowchart summarising \textit{EfficientGrasp} is shown in Fig.~\ref{fig:flowchart}. The main contribution of this paper is that compared to the existed method, the proposed method is more data-efficient, accurate, and can generate more reliant grasps for different grippers.
 
 The benefits of the proposed method are, firstly, that using fingertip workspaces is more data-efficient than using a full model of the gripper itself (with an average reduction in memory use of 81.7\% according to our experiments). Secondly, that the proposed method is more generalizable to novel grippers as it does not rely on model specifications such as a URDF, which may have problems when representing gripper geometries based on closed-loop constraints. Indeed, our real-world experiments, conducted with the RUTH hand~\cite{luruth}---a gripper with closed-loop constraints, show that UniGrasp fails while \textit{EfficientGrasp} achieves a grasping success rate of 83.3\%. Thirdly, that the proposed method also outperforms the baseline when considering only grippers without closed-loop constraints (e.g., Robotiq-3F gripper, BarrettHand), showing 9.85\% higher accuracy in contact points generation compared to UniGrasp and 3.10\% higher grasping success
rate in simulation (86.3\% \textit{EfficientGrasp}, 83.2\% UniGrasp), which indicates that the proposed method can generate better grasps.


\section{Related Work}
\label{section2}

\subsection{Grasping Using Different Grippers}
The most successful grasp synthesis approaches use data-driven methods which can generalize to new, unseen objects \cite{joshi2020robotic,ni2020pointnet++,liang2019pointnetgpd,boroushaki2021robotic,bousmalis2018using}. These approaches commonly use a parallel jaw gripper attached to the end of a robot arm. Parallel jaw grippers are simple systems with minimal degrees of freedom and therefore the process of generating grasping configurations is made more straightforward, for example by producing a grasp rectangle~\cite{guo2017hybrid}. This limits the grasp synthesis to only two contact points, and thus the number of grasp types that can be achieved is restricted.

In recent years, several grasping methods using fully-actuated multi-finger robot hands have been proposed \cite{fan2018real,hang2016hierarchical,lundell2021multi}, increasing the number of contact points and thus increasing the potential number of objects that can be successfully grasped. 
Fully-actuated grippers require large numbers of actuators to achieve a wide range of grasp configurations, but this means they are more vulnerable to the accumulation of actuator errors, limiting the accuracy that these hands can achieve, and thus complex algorithms are, in general, required to control them. 

Underactuated hands have been proposed as an alternative to achieve dexterity with a reduced number of actuators. Multiple finger phalanges are controlled by a single motor, which is often achieved through the use of tendon driven systems and joint coupling\cite{hang2020hand,lu2019soft,kwon2021underactuated}. This simplifies the control of the gripper and reduces the effects of actuator error accumulation, however with limitations in performing precision grasps.
Multi-finger robot gripper designs with a reconfigurable palm can achieve an even wider range of grasp configurations with high precision
\cite{luruth,lu2021systematic,spiliotopoulos2018reconfigurable,borisov2021computational}. The reconfigurable palm design allows for finger positions and orientations to be controlled relative to each other. This type of hand, along with others that are able to generate multiple types of grasps, are further discussed in section \ref{sec:multigrasp}.

\subsection{Visual Data-driven Grasping Methods}
Data-driven grasping approaches have gained popularity in recent years, and have proven to be particularly useful when applied to novel objects~\cite{goldfeder2011data}. These methods vary in terms of the nature of the input data they use and how the grasp is represented in the output. Deep learning is one of the most popular data-driven methods. Some have used Convolutional Neural Networks (CNNs) to detect objects from an RGB-D camera image and generate valid grasp locations for a specific object~\cite{pinto2016supersizing,lenz2015deep}. CNNs can also be applied for real-time grasp detection~\cite{redmon2015real}, which expands the potential for applying CNN-based visual grasping in daily life. A novel grasping method that can predict a score for every possible grasp pose is presented in \cite{johns2016deep}. This method makes it possible to achieve grasping robust to the gripper's pose uncertainty.


Many works have focused on generating large datasets which can be leveraged to learn better quality grasps. The \textit{Dexnet} dataset contains 10,000 3D object models~\cite{mahler2016dex}, which was also used to generate a larger dataset of synthetic object points~\cite{mahler2017dex}. This dataset was used to train a Grasp Quality Convolutional Neural Network (GQ-CNN) which predicts the grasp quality of parallel-jaw grasps, and this was later extended to suction grippers~\cite{mahler2018dex} and ambidexterous grasping~\cite{mahler2019learning}.

The limitation of all of these methods is that they are trained for a single gripper. The UniGrasp framework~\cite{shao2020UniGrasp} addresses this shortcoming by presenting a grasp generative approach which is generalizable not only to novel objects, but novel grippers as well---provided that the gripper does not have closed-loop kinematic chains. This method is based on~\cite{qi2017pointnet++} to extract the features of grippers and objects from their point clouds, such that it can train a Point Set Selection Network (PSSN), which can generate contact points on any object for any gripper. However, the method becomes data-inefficient and cumbersome when applied to multi-grasp grippers such as the Robotiq-3F gripper, the BarrettHand, and the RUTH hand~\cite{luruth}; and does not work on grippers with closed-loop constraints (e.g., the RUTH hand, the integrated linkage-driven hand~\cite{kim2021integrated}). These drawbacks are discussed in section \ref{sec:multigrasp}.

\subsection{Grasping with Multi-grasp Robotic Hands} \label{sec:multigrasp}
Multi-grasp robotic hands can be defined as grippers with three or more fingers whose proximal joints are not fixed and can be repositioned. This feature brings significant advantages, including in-hand manipulation capabilities and the ability to achieve a wider range of grasping configurations. Multi-grasp grippers are widely used due to their advantages in manipulation and grasping; examples include the Robotiq-3F gripper, the BarrettHand, and the RUTH hand~\cite{luruth}. The first two are fully-actuated, while the RUTH hand has underactuated fingers and a palm based on a five-bar linkage.



UniGrasp~\cite{shao2020UniGrasp} is a unified method that can be generalized to both novel objects and novel grippers. UniGrasp constructs 'gripper features' by taking point clouds of the gripper in each operation mode for which each actuated degree of freedom is in its most `open' and `closed' position. A large number of samples is required for grippers with multi-grasp configurations. In the case of a multi-grasp gripper such as the RUTH hand~\cite{luruth}, the finger base positions and orientations are able to move anywhere along the 2-dimensional manifold corresponding to the set of configurations the five-bar linkage can achieve.

\begin{figure}[t]
    \centering
    \includegraphics[width=0.7\linewidth]{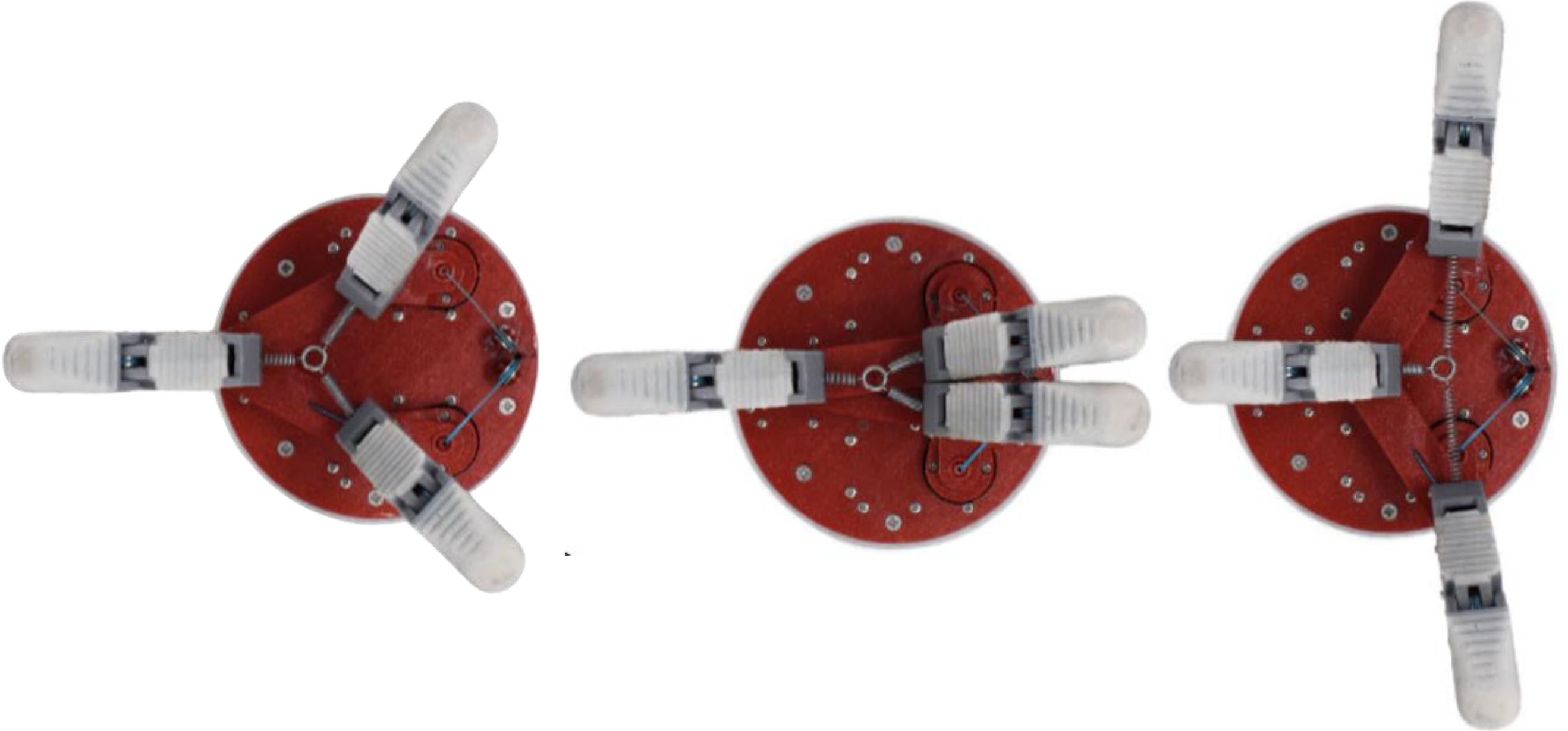}
    \caption{Example of a multi-grasp gripper (the RUTH hand) in different operation modes.}
    \label{fig:GraspTypes}
\end{figure}

Therefore, the most `open', `closed' and `average' operation modes of a multi-grasp gripper are ambiguous, each of which may be most suitable for performing different grasp types, as is shown in Fig.~\ref{fig:GraspTypes}. Indeed, UniGrasp takes 32 different operation modes for the Robotiq-3F gripper, and 256 different operation modes for the BarrettHand. This implies that exhaustive operation mode data generation and collection is required when the method is applied to multi-grasp grippers.

Once the contact points are generated, the inverse kinematics solution for the gripper needs to be computed. However, current available open source libraries for solving robot kinematic problems such as RBDL~\cite{felis2017rbdl} and IKPy always require the gripper URDF model as input. This is infeasible for closed-loop grippers such as the RUTH hand, as its reconfigurable palm utilizes a closed-loop five-bar linkage mechanism controlled by two actuated joints, while URDF does not permit such closed-loop description. 

Our work aims to address these drawbacks in the current grasping methods and IK approaches, to develop a method of generative grasping which is applicable to reconfigurable grippers, as well as conventional grippers. In this paper, we use the fingertip workspace of the gripper to generate the corresponding gripper feature, since the workspace of a gripper is unique and can be clearly defined, such that the configuration implicity problem is overcome. We then use a model-free method to solve the inverse kinematics of the gripper, such that URDF models are no longer required.

\section{Grasping Method}
\label{section4}

\subsection{Extracting Gripper Features}
To train the PSSN model based on the gripper fingertip workspace, the global features of the workspace should be extracted beforehand. 
\begin{figure}[t]
    \centering
    \includegraphics[width=0.9\linewidth]{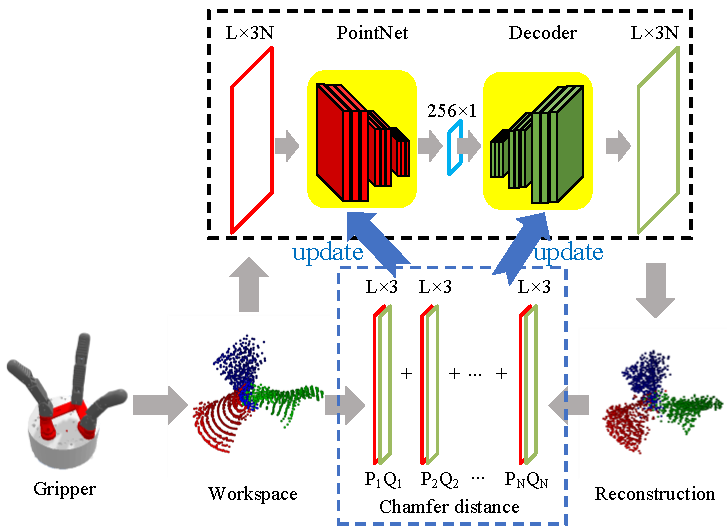}
    \caption{PointNet is trained along with a decoder for extracting features based on Chamfer distance.}
    \label{fig:autoencoder}
\end{figure}
For an $N$-fingered gripper, its fingertip workspace has a dimension of $L\times3N $ where $ L $ denotes the number of workspace point sets. Each row of the workspace matrix represents a set of fingertip positions. The fingertip workspace $ S\in \mathbb{R}^3$ can also be represented as
 \begin{equation}
 S=
 \left[
 \begin{array}{cccc}
     P_{11} & P_{12} & \cdots & P_{1N} \\
     P_{21} & P_{22} & \cdots & P_{2N} \\
     \vdots & \vdots & \ddots & \vdots \\
     P_{L1} & P_{L2} & \cdots & P_{LN}
 \end{array}
 \right]    
 \end{equation}
 where $ P_{mn} $ with $ m=1,2,\cdots,L,\,n=1,2,\cdots,N $ is a $ 1\times3 $ vector representing a point in the fingertip workspace. By transforming the workspace into a point cloud, the global features can be extracted using PointNet~\cite{qi2017pointnet}, which has shown good performance in processing point clouds. Inspired by PointNet, our network has 6 convolutional layers followed by 1 max pooling layer. All of the 6 convolutional layers are activated by ReLU functions. Through this modified PointNet architecture, the input of an $ L\times3N $ fingertip workspace will be reduced to the $ 1\times256 $ latent feature space. To train the PointNet, we use a decoder with 3 fully connected layers to reconstruct the $ L\times3N $ fingertip workspace. The first 2 layers are activated by ReLU functions while the last layer is activated by a linear function. Then we train the PointNet and decoder, following an autoencoder structure, by minimizing the Chamfer distance between the input and reconstructed point cloud.

Unlike the traditional Chamfer distance method, there are $N$ coupled points in each row of the fingertip workspace point cloud. We define the coupled Chamfer distance between $  S_1,S_2\in \mathbb{R}^3 $ as following:
\begin{equation}
\begin{split}
d_{CD}(S_1,S_2) = \sum_{i=1}^N(\sum_{j=1}^L\min_k\Vert P_{j,i}-Q_{k,i} \Vert_2^2\\
+\sum_{j=1}^L\min_k\Vert Q_{j,i}-P_{k,i} \Vert_2^2)
\end{split}
\end{equation}
where $ P \in S_1 $ and $ Q \in S_2 $ respectively. Fig.~\ref{fig:autoencoder} shows the PointNet and decoder training process in detail.

\subsection{Generating Contact Points}

\begin{figure}[t]
    \centering
    \includegraphics[width=0.95\linewidth]{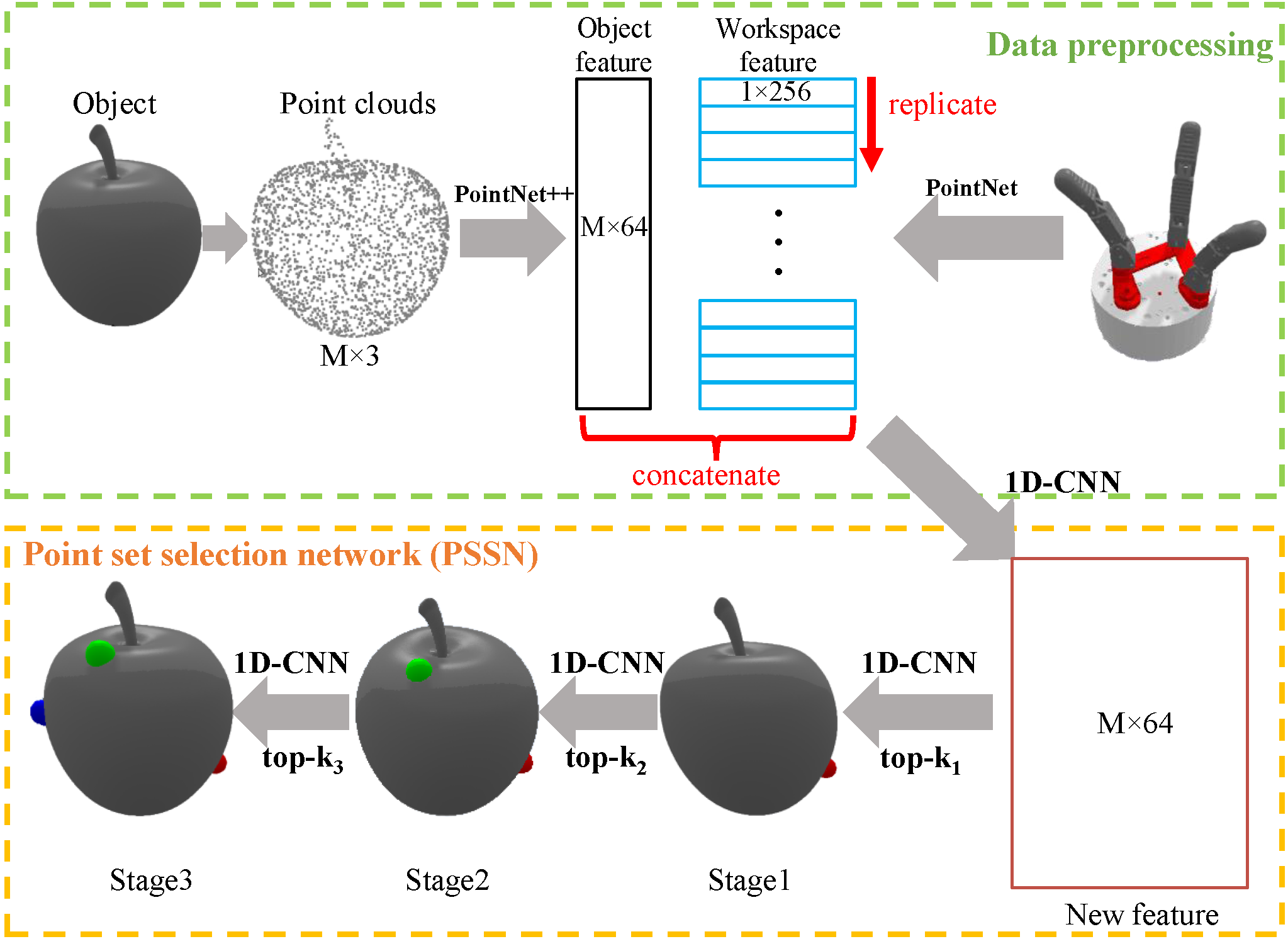}
    \caption{Preprocessing the extracted features and feeding them into a trained PSSN to generate the contact points.}
    \label{fig:PSSN}
\end{figure}

The gripper's global feature is preprocessed for generating the contact points for a specific object and gripper as shown in Fig.~\ref{fig:PSSN}. First, the fingertip workspace feature ($ 1\times256$) can be extracted by feeding its workspace point cloud into the trained PointNet. Next, an object point cloud ($ M\times3 $) is generated from its OBJ format file, where $ M $ represents the number of points in the point cloud. Then we use PointNet++~\cite{qi2017pointnet++} to extract the $ M\times64 $ object feature from the generated object point cloud. Every line of the object feature represents the feature of each point in the object point cloud. In order to create a connection between the object and gripper to evaluate whether the points in the object point cloud are reachable by the gripper, we replicate the gripper feature along the point's dimension for $ M $ times to concatenate the two features. Finally, these features are concatenated and a 1D-CNN is used to acquire a compact, lower dimension, new feature, which will be supplied as the input for generating the contact points.

Inspired by the \textit{UniGrasp} framework, we apply similar structure of the PSSN to find valid contact points on the surface of the object one by one. The PSSN works as follows: In stage one, the $ M\times64 $ feature is fed into a 1D-CNN and a softmax layer to calculate the probability of being a valid point for each point from the point cloud. The points are then ranked and the top $ k_1 $ points are selected for stage two. The other contact points will be generated in the next stages.

The workspace-based PSSN and the original PSSN are both trained in this paper. We select $k_1=1024$ in stage one, $k_2=512$, and $k_3=512$ in stage two, and stage three, respectively. The length of the object point cloud is set as $ M=2048 $. We apply an Adam optimizer with a learning rate of $2e^{-3}$ to train the PSSN. The networks in each stage are trained one by one: In the first stage of training, only the loss of stage one will be calculated and only the weights of this stage will be updated. During stage two of training, the weights of stage one are fixed and only the loss and weights of stage two will be computed. Finally we repeat the same operation for stage three. The datasets for training and testing in this paper are split from the full datasets of~\cite{shao2020UniGrasp}. The size of the used training data and test data are 600 GB and 150 GB respectively. Each of the stages are trained for 220 epochs. All of the training is performed on a PC with an Intel Core i7-9700K CPU and NVIDIA RTX 2080 Ti.

\subsection{Computing Gripper Inverse Kinematics}

 Model-free reinforcement learning (RL) has shown advantages in training stability in solving IK for robot arms with continuous action space. When the robot arm is attached with a gripper with multiple fingers, the states and actions of the combined robotic arm-hand system have significantly higher dimensions and such a system can suffer from the curse of dimensionality when training an RL policy. To maximize training efficiency, grasp motion is divided into two steps: First, compute a rest position and orientation in which the robot's end effector is a distance $d$ away and perpendicular to the plane of the contact points. Next, manoeuvre the gripper configuration to best reach the contact points.
\begin{figure}[t]
    \centering
    \includegraphics[width=0.7\columnwidth]{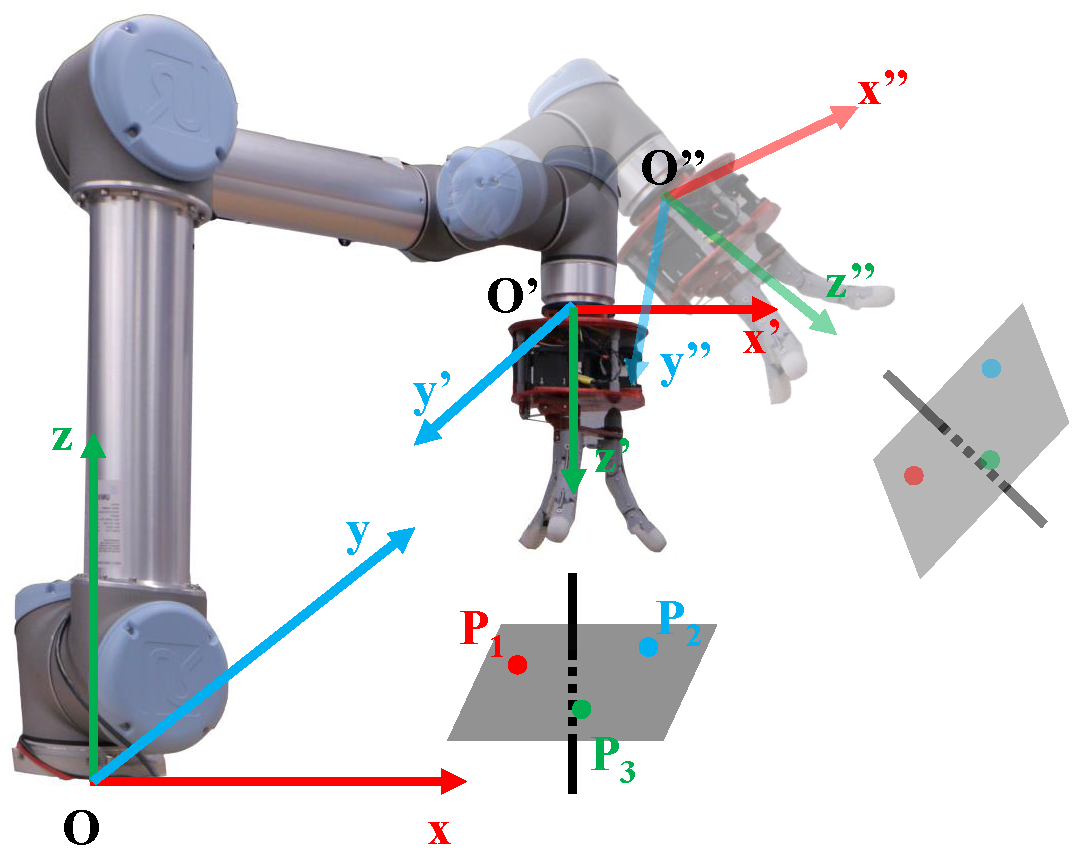}
    \caption{Visualisation of the world frame to end-effector frame transformation. Contact points $P_{1,2,3}$ retain the same coordinates in the end-effector frame regardless of end-effector orientation.}
    \label{fig:RL}
\end{figure}

The policy is trained at a fixed end-effector orientation under the end-effector frame. This means the trained policy is independent of the orientation of the end effector and can be generalized to any robot arm configurations. As shown in Fig.~\ref{fig:RL}, contact points $P_1, P_2, P_3$ have the same coordinates using the end-effector frame even if the robot arm configuration changes.
The target points are transformed from the world frame to the end-effector frame using the transformation,
\begin{equation}
^eP^g=^eT_w\cdot ^wP^g
\end{equation}
where $^eT_w$ is a homogeneous transformation matrix which maps the points in the world frame, $^wP^g$, to the points in the end-effector's frame, $^eP^g$.

In this work, soft actor critic (SAC)~\cite{haarnoja2018soft} method is implemented to train a stochastic policy for changing the gripper configuration, which outputs motor values to move the fingertip positions $P^o$ into the target positions $P^g$ ($P^o\neq P^g$). The agent observes a state $s_t$ (including fingertip positions, joint values) and executes an action $a_t$ (gripper motor values) sampled from the policy. We append the previous action to the inputs to the policy (i.e., $a_t=\pi(s_t,a_{t-1})$) to make the policy smooth. The policy aims to maximize the reward, which is computed based on the square sum of final fingertip position error of all fingers, defined as:
\begin{equation}
r=-\sum_{i=1}^N\Vert ^wP_i^o-^wP_i^g \Vert_2^2
\end{equation}
where $N$ refers to the number of fingers. For the RUTH hand, we define the action space as $\mathbb{A}^4$, including the actuated joint positions of the gripper and the last joint of the robot arm. The observation space is defined as $\mathbb{S}^{13}$, which includes the joints and the coordinates of the target fingertip positions.
\begin{table}[t]
\centering
\begin{threeparttable}
\caption{Summary of Contact Points Generation Average Performance}
\begin{tabular}{cc|c|c}
\hline
\multicolumn{2}{c|}{}                                                                                                & UniGrasp   & EfficientGrasp (Ours)         \\ \hline
\multicolumn{2}{c|}{Input Data}                                                                                      & Gripper PC & Workspace PC \\ \hline
\multicolumn{2}{c|}{Input Number}                                                                                    & $\geq$3          & 1            \\ \hline
\multicolumn{2}{c|}{\begin{tabular}[c]{@{}c@{}}Feature Extraction \\ Time (s)\end{tabular}}                          & 2.04       & 1.67         \\ \hline
\multicolumn{2}{c|}{Memory Usage (MB)}                                                                               & 4777       & 873          \\ \hline
\multicolumn{1}{c|}{\multirow{3}{*}{\begin{tabular}[c]{@{}c@{}}PSSN \\ Training \\ Time (s)\end{tabular}}} & Stage 1 & 31.43      & 30.16        \\ \cline{2-4} 
\multicolumn{1}{c|}{}                                                                                      & Stage 2 & 313.67     & 311.83       \\ \cline{2-4} 
\multicolumn{1}{c|}{}                                                                                      & Stage 3 & 550.81     & 542.91       \\ \hline
\multicolumn{2}{c|}{PSSN Size (MB)}                                                                                  & 316.8      & 282.2        \\ \hline
\end{tabular}
\label{table:efficiency}
\begin{tablenotes}   
        \footnotesize    
        \item Extracting gripper features using UniGrasp and our method, tested on Robotiq-3F, Kinova and BarrettHand. The RUTH hand is not included as UniGrasp cannot be applied to it. Data in the table are average values.
      \end{tablenotes}           
\end{threeparttable}
\end{table}

\section{Method Evaluation}
\label{section5}

\subsection{Contact Points Generation Performance}

\begin{table}[t]
\centering
\begin{threeparttable}
\caption{PSSN Accuracy}
\begin{tabular}{c|r|r|r|r|r|r}   
\hline   
\multirow{2}*{}&
\multicolumn{2}{c|}{Stage One ($\%)$} & \multicolumn{2}{c|}{Stage Two ($\%)$} & \multicolumn{2}{c}{Stage Three ($\%)$}\\
\cline{2-7} 
&Top1& Top10& Top1& Top10& Top1& Top10\\
\hline
\textit{\textbf{E}} Robotiq-3F & \textbf{99.3}& \textbf{100}& \textbf{97.5}& \textbf{98.9}& \textbf{88.7}& \textbf{96.1}  \\ 
\textit{\textbf{U}} Robotiq-3F & 91.9& 99.6& 83.4& 90.8& 76.0& 86.2  \\ \hline
\textit{\textbf{E}} Kinova-3F & \textbf{97.9}& \textbf{100}& \textbf{87.6}& \textbf{97.2}& \textbf{75.6}& \textbf{89.8}  \\ 
\textit{\textbf{U}} Kinova-3F & 86.9& 98.2& 66.4& 80.2& 54.8& 70.7  \\ \hline
\textit{\textbf{E}} BarrettHand & \textbf{100}& \textbf{100}& \textbf{96.5}& \textbf{99.7}& \textbf{91.9}& \textbf{96.5}  \\ 
\textit{\textbf{U}} BarrettHand & 94.3& 98.9& 86.9& 91.5& 77.7& 87.6  \\ 
\textit{\textbf{I}} BarrettHand & 92.2& 98.6& 85.9& 90.8& 77.4& 86.9  \\
\hline   
\end{tabular} 
\label{table:accuracy}
\begin{tablenotes}   
        \footnotesize    
        \item Each stage of PSSN is trained for 220 epochs on 600 GB datasets, tested on 150 GB datasets. The prefix \textit{\textbf{E}} represents tests using EfficientGrasp, while \textit{\textbf{U}} means tested on original UniGrasp. \textit{\textbf{I}} means extracting gripper feature using insufficient inputs for UniGrasp.
      \end{tablenotes}           
\end{threeparttable}
\end{table}

\begin{figure*}
\centering    
\subfigure[Stage 1]{\label{fig:a}\includegraphics[width=0.31\textwidth]{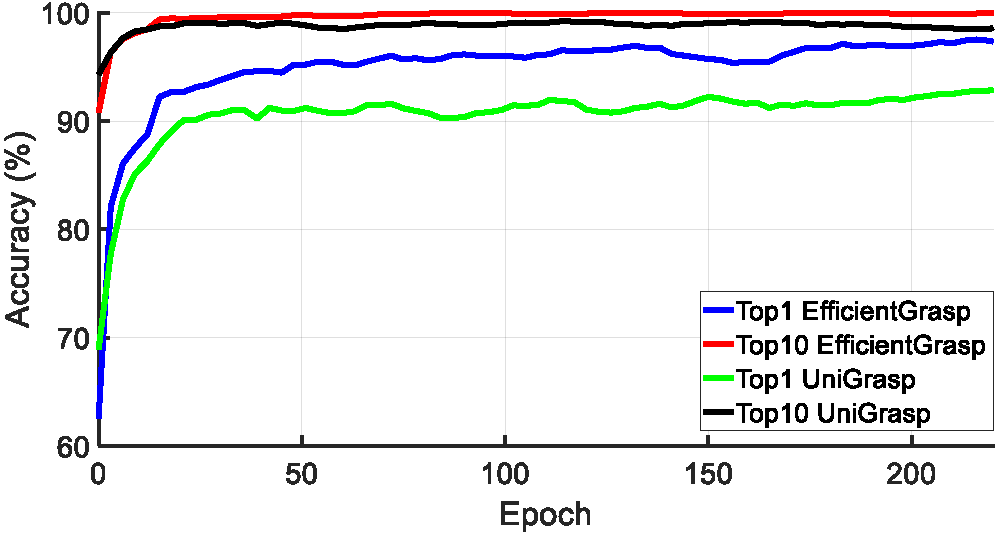}}
\hspace{0.02\textwidth}
\subfigure[Stage 2]{\label{fig:b}\includegraphics[width=0.31\textwidth]{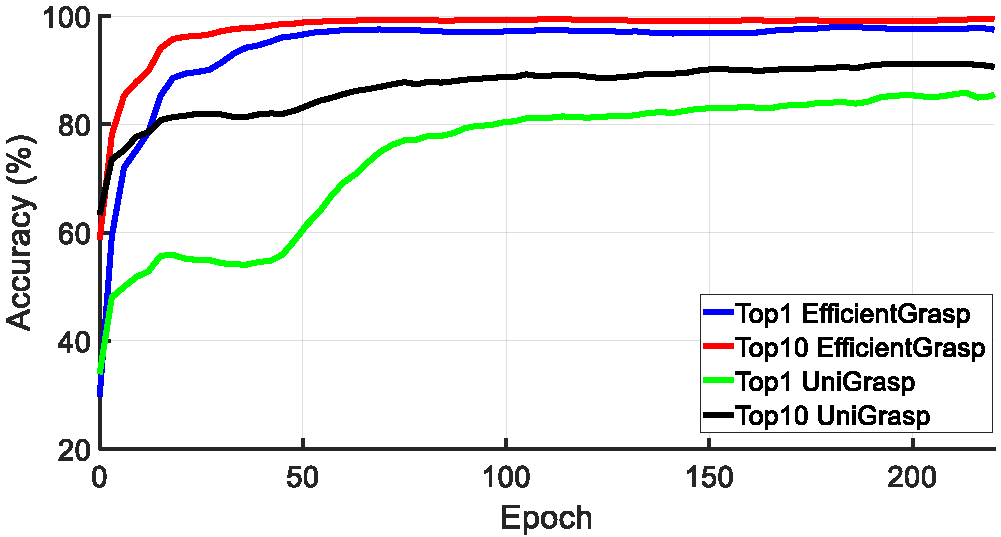}}
\hspace{0.02\textwidth}
\subfigure[Stage 3]{\label{fig:b}\includegraphics[width=0.31\textwidth]{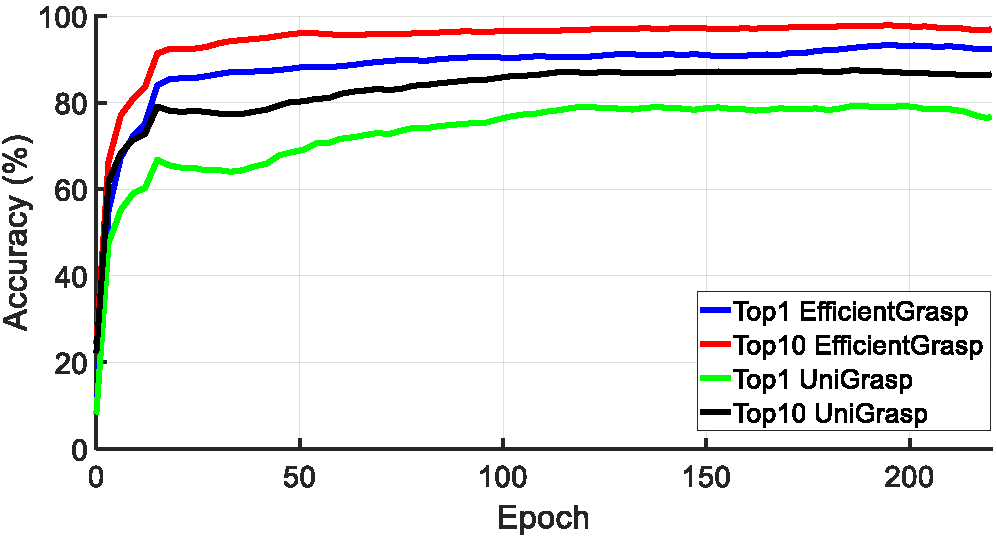}}
\caption{Comparison of our PSSN accuracy with UniGrasp accuracy at each training stage (trained with BarrettHand).}
\label{fig:accuracy}
\end{figure*}

In this section, we compare the contact points generation performance of the proposed method with UniGrasp. The RUTH hand is not included for this evaluation because UniGrasp does not work for it.  The input number for UniGrasp is different depending on the gripper type. As shown in Table~\ref{table:efficiency}, while only 1 PC from each gripper is required for \textit{EfficientGrasp}, UniGrasp takes 32, 8, and 254 operation modes as inputs for Robotiq-3F, Kinova-3F, and BarrettHand, respectively. Therefore, our method shows better data-efficiency and generalization compared to the UniGrasp framework.

Next, we train the PSSN using the gripper features generated by our method and UniGrasp. With the trained PSSNs, we test their accuracy on the above grippers. We evaluate the performance of the PSSNs by calculating the accuracy of the highest ranked point to be within 5mm of a ground truth set of points, which is `Top1', while `Top10' describes the percentage of tests for which at least one of the ten highest ranked points is within 5mm of the ground truth point set.

As shown in Table~\ref{table:accuracy}, the proposed method outperforms the original UniGrasp for all of the tested grippers. Importantly, the performance of UniGrasp can be affected if the input gripper configurations are insufficient. Our method avoids this problem by using the fingertip workspace, since it is unique for each specific gripper. The accuracy for the BarrettHand in each training epoch is shown in Fig.~\ref{fig:accuracy}. Our method outperforms the original UniGrasp in all three stages after training for a small number of epochs. The UniGrasp algorithm requires more training epochs (about 100 epochs) to converge.

\subsection{Model-free IK Performance}

The proposed model-free RL method for IK is evaluated in this subsection. We utilize OpenAI GYM to set up the simulation environment. The action and observation spaces are defined as described in~\ref{section4}. During training, the orientation of the UR5 end effector is fixed. In each epoch, 10 sets of target points are generated randomly from the gripper workspace. For each target point, the RL policy runs for up to 100 episodes. As for the parameters of SAC, we choose Gaussian policy and set the temperature parameter as $\alpha=0.8$, the learning rate as $lr=0.003$, the target smoothing coefficient as $\tau=0.005$, and the reward discount factor as $\gamma=0.99$. The device for training the RL policy is a NVIDIA GeForce RTX 3070.
\begin{figure}[tbp]
    \centering
    \includegraphics[width=1\columnwidth]{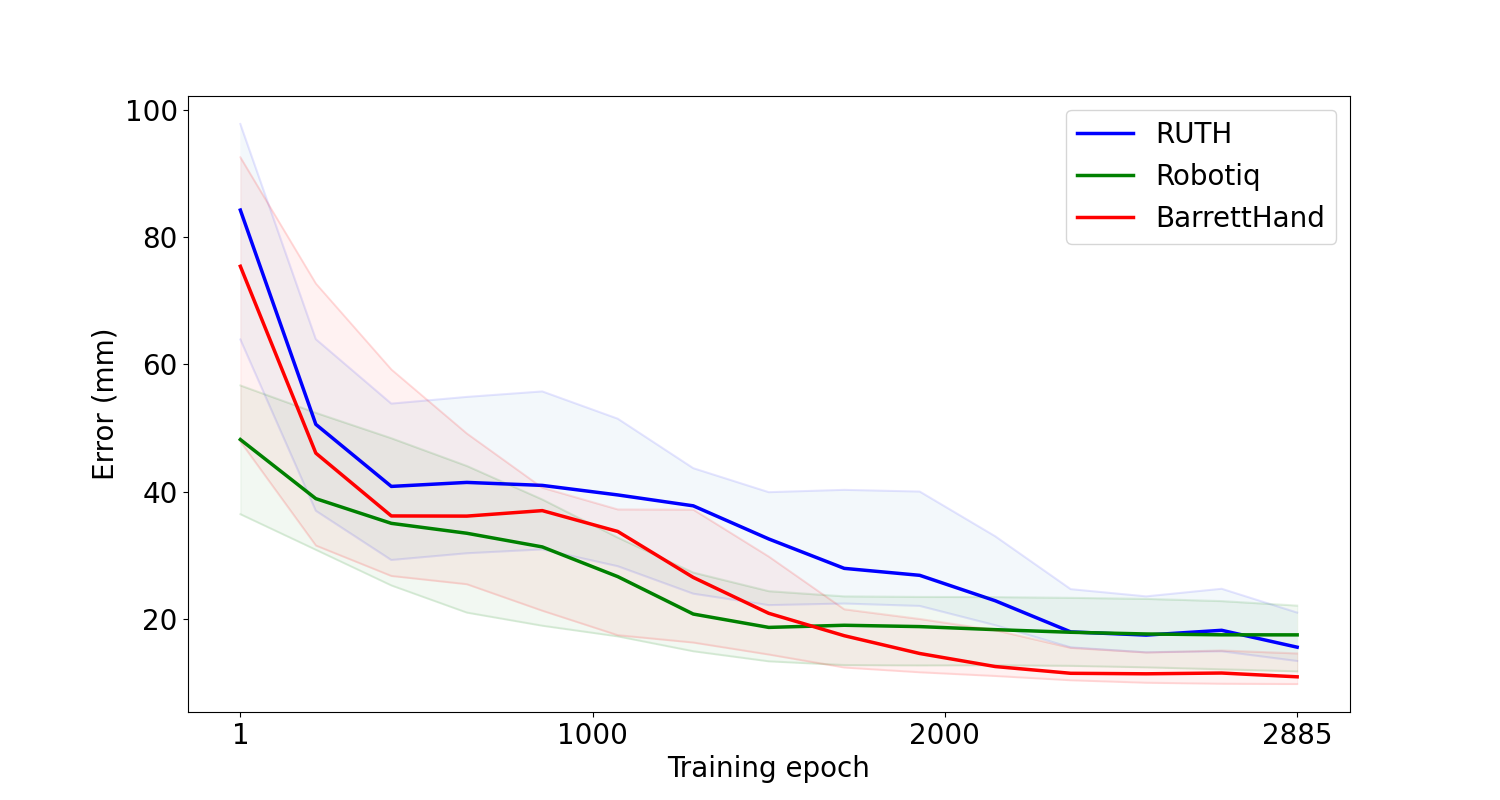}
    \caption{Plot of model-free IK RL evaluation performance, showing the error range at epoch 1, 100, 1000 and 2885 (convergence). 100 sets of contact points were generated for each epoch.}
    \label{fig:RL_eval}
\end{figure}

In order to show the feasibility and effectiveness of the proposed model-free RL method, we implement it to two commercial grippers: Robotiq-3F and BarrettHand BH8-280, and a RUTH gripper, which has closed-loop chains. The sum of distance error between each fingertip positions and their corresponding target positions are computed and recorded throughout the RL training in Fig. \ref{fig:RL_eval}. After 2885 epochs of training, the average error is reduced by 81\% to within 5mm per finger for all three grippers. These results reveal that the proposed RL method is effective for solving IK for a wide range of novel grippers without the knowledge of the gripper's mathematical model.

\section{Object Grasping}
\label{section6}

\subsection{Grasp Quality}
\label{subsection61:GQS}
\begin{figure}[t]
    \centering
    \subfigure[]{
	\label{fig:quality:a} 
	\includegraphics[width=0.35\columnwidth]{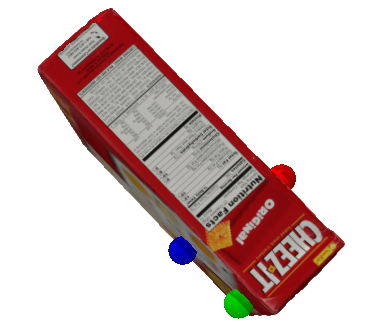}}
    \subfigure[]{
	\label{fig:quality:b}
	\includegraphics[width=0.35\columnwidth]{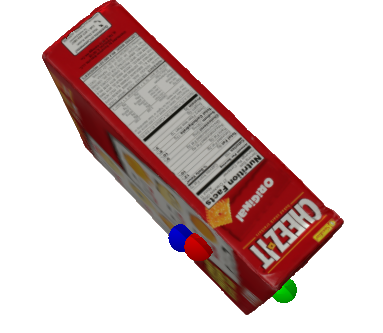}}
	\subfigure[]{
	\label{fig:quality:c} 
	\includegraphics[width=0.35\columnwidth]{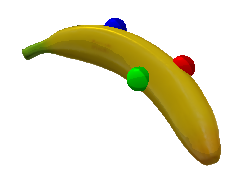}}
    \subfigure[]{
	\label{fig:quality:d}
	\includegraphics[width=0.35\columnwidth]{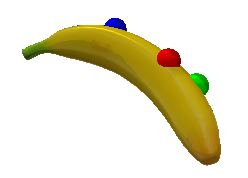}}
\caption{Examples of different sets of contact points on objects. (a) GQS=0.9906 and (c) GQS=0.9157 indicate these contact points are in force closure condition. (b) GQS=0.5591 and (d) GQS=0.4649 correspond to grasps that fail to meet the force closure condition.}
\label{fig:quality}
\end{figure}

We calculate a Grasp Quality Score (GQS) to evaluate the quality of contact points generated from the PSSN. The GQS is a metric adopted from the force closure estimator \cite{liu2021synthesizing}, reflecting how well a set of contact points satisfies force closure condition. A force-closure grasp is defined as a grasp that can resist arbitrary external wrenches while contact forces are applied within the friction cones rooted from the contact points. Examples are shown in Fig.~\ref{fig:quality}. Contact point sets with GQS ($\in [0,1]$) higher than 0.75 are considered valid and are expected to generate force-closure grasps. 
GQS is computed using the following formula:
\begin{equation} \label{GQS}
GQS=2-\frac{2}{1+exp(-\lambda_0(GG'-\epsilon I_{6\times6}))},
\end{equation}
where $G$ is a matrix containing the locations of the contact points and $\lambda_0(\cdot)$ gives the smallest eigenvalue.  

Our trained PSSN returns a list of contact point sets ranked by the highest GQS. 
In the following section, we evaluate and validate the generated contact points by performing grasping tests on YCB objects~\cite{Xiang-RSS-18} both in simulation and in real-world experiments. The inverse kinematics of the tested multi-finger grippers is solved using their corresponding trained RL policy, such that the fingertips match the positions of the specified contact points during grasping.

\subsection{PyBullet Simulation}
Three multi-finger grippers are simulated and tested in PyBullet environment: Robotiq-3F, BarrettHand BH8-280 and RUTH hand. In each test, the selected gripper is attached to the end effector of a UR5 robot arm. We test each gripper system with 16 YCB objects. Each object is tested 5 times, placed in random orientations and locations (within the robot arm workspace and in the view range of the synthetic camera). The PSSN returns a best set of contact points using the gripper workspace feature and camera-captured point cloud as inputs. Finally, the trained RL policy is used to solve for IK and outputs the motor control for each gripper.
Table~\ref{table:grasp} shows that our proposed method achieved a grasp success rate of 85.0\%, 87.5\% and 83.8\% for Robotiq, BarrettHand and RUTH grippers, respectively, across all object trials in simulation. The average success rate is 85.4\% across all gripper trials. This is higher than the UniGrasp framework, which achieved a success rate of 81.3\% and 85.0\% for Robotiq and BarrettHand, with 0\% for the RUTH hand as the method fails in this case.

\begin{table*}[t]
\centering
\begin{threeparttable}
\caption{Object Grasping Simulation and Real-World Experiment Results}
\begin{tabular}{l c c c | c c |c c}
\hline
& & & & \multicolumn{2}{c|}{Baseline (UniGrasp)} & \multicolumn{2}{c}{EfficientGrasp} \\
Gripper & Objects & Trials per object & Total trials & Success Rate (\%) & Mean GQS & Success Rate (\%) & Mean GQS \\ \hline
Robotiq-3F (S) & 16 & 5 & 80 & 81.3 &  0.8188    & {\bf 85.0}              & 0.8387    \\         
BarrettHand (S) & 16 & 5  & 80 & 85.0 & 0.8170     & {\bf 87.5}              &  0.8518    \\         
RUTH (S) & 16 & 5  & 80 & Failed (0) & --   & {\bf 83.8}              & 0.8587            \\ 
RUTH (RW) & 18 & 5    & 90  & Failed (0) & --  & {\bf 83.3}              & 0.8868            \\ \hline
\end{tabular}
\label{table:grasp}
\begin{tablenotes}   
        \footnotesize    
        \item  Each object was attempted 5 times,
placed in random orientations and locations within the robot arm workspace and in the view range of the camera. A successful grasp is recorded when the robot grasps the
object at the specified contact points and holds the object for 5 seconds above the table. A Grasp Quality Score (GQS) $\in [0,1]$ higher than $0.75$ is considered to generate force-closure grasps. (S) denotes simulation and (RW) real world.
      \end{tablenotes}
\end{threeparttable}
\end{table*}

\subsection{Real-World Experiments}

\begin{figure}[t]
    \centering
    \includegraphics[width=0.7\columnwidth]{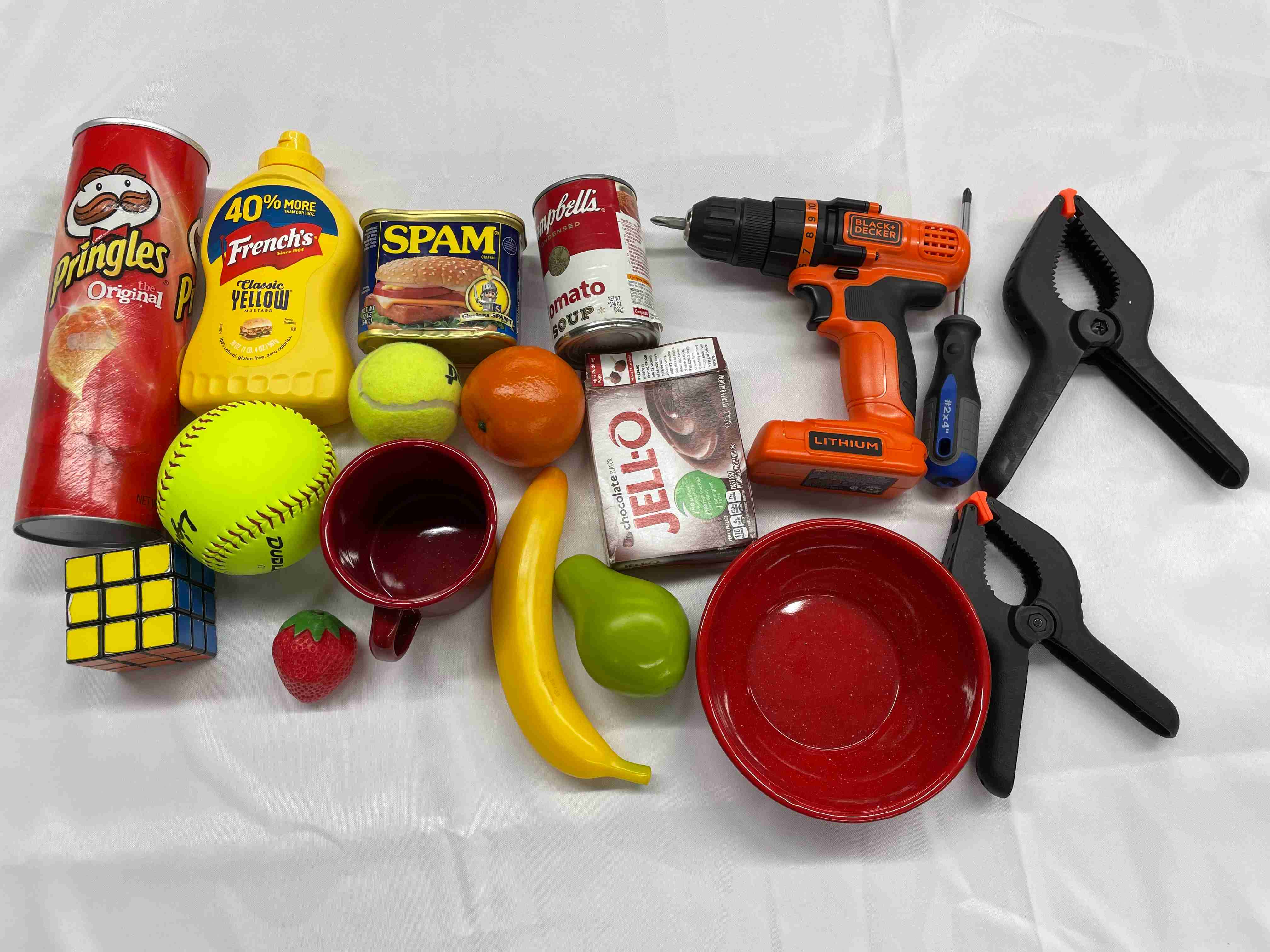}
    \caption{18 YCB objects were tested on a real robotic manipulator. EfficientGrasp method achieves a success rate of 83.3\%.}
    \label{fig:YCB}
\end{figure}

The proposed method is tested in reality using a UR5 robot arm attached with the RUTH hand to grasp 18 objects from the YCB dataset (Fig. \ref{fig:YCB}). Test protocol is kept the same as in simulation. Each object is tested 5 times, placed at random locations within the robot arm workspace in random orientations in each trial. A successful grasp is recorded when the robot grasps the object at the specified contact points and holds the object for 5 seconds above the table.  A Realsense D435i camera is placed in the same location and orientation as in simulation, to capture a depth image of the object for point cloud generation. The RUTH gripper workspace feature and captured point cloud are used as inputs for the PSSN to generate valid contact point sets. The trained RL policy is implemented while grasping.

Our method achieves an overall success rate of 83.3\% across a total of 90 grasp trials on 18 objects in real world. This is similar to the grasp success rate of 83.8\% for the simulated RUTH-UR5 robot system. The real world experiments prove that the contact points generated by our PSSN indeed satisfy force-closure, hence validates the application of the \textit{EfficientGrasp} method in contact point generation and in computing model-free gripper inverse kinematics. 

\begin{figure}[t]
    \centering
    \subfigure[]{
	\label{fig:RUTH:a} 
	\includegraphics[width=0.19\columnwidth]{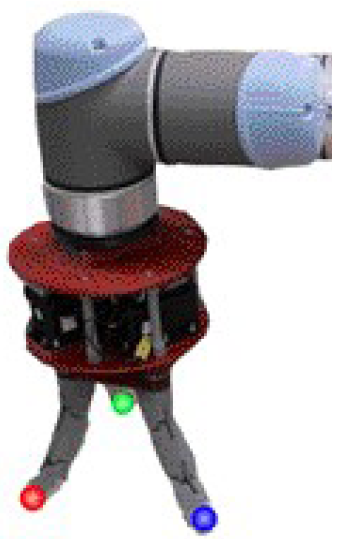}}
    \subfigure[]{
	\label{fig:RUTH:b}
	\includegraphics[width=0.18\columnwidth]{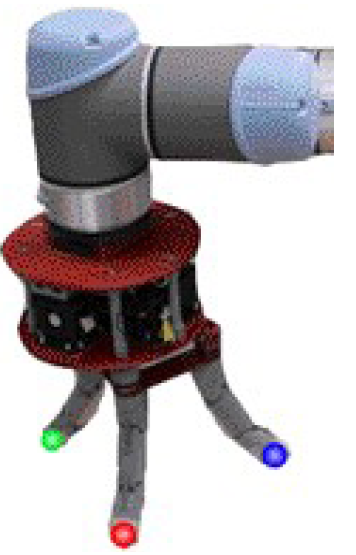}}
	\subfigure[]{
	\label{fig:RUTH:c} 
	\includegraphics[width=0.2\columnwidth]{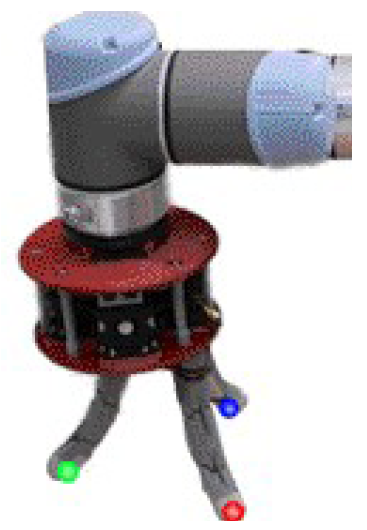}}
\caption{Example of the RUTH hand configuring from initial position (a) to end position (c) in a real experiment. Full experiment video available in the supplementary material.}
\label{fig:RUTH-real}
\end{figure}

\subsection{Failure Analysis of Real-World Experiments} \label{sec: failure analysis}
We evaluated and compared the quality of contact points generated from synthetic and real-life object point clouds using the Grasping Quality Score (GQS)  \cite{liu2021synthesizing}. The GQS of the best contact points generated for each object trial in simulation (80 trials) and real world (90 trials) was computed with a mean of 0.8587 and 0.8868, respectively. This indicates a high confidence in forming force-closure grasps ($>0.75$) for both cases. With the contact points generated, the data-driven inverse kinematics implemented for the RUTH gripper was able to position the fingertips to reach them satisfactorily. 



Although all selected contact points satisfied force closure condition, some of these point sets were out of reach of the robot arm in reality. This is most prevalent when an object is placed in a highly self-occluded orientation with respect to the camera viewpoint. The captured point clouds results in contact points that encourage the robot to approach from the side of the object that is closer to the camera but further from the base of the robot arm, which resulted in unreachable targets and failed grasps. Moreover, some objects have different physical properties in reality which were not accounted for in simulation. For example, in real world experiments, the gripper was able to grasp metal objects and lift it slightly, but was not able to hold it for 5 seconds as the object slips easily due to low surface friction. 

All in all, the failed grasp trials in real experiments are mainly caused by the difference in physical properties of the object in reality (e.g. surface friction, weight) and occlusions in captured object point clouds which resulted in unreachable contact points by the robot arm.


\section{Conclusion}
\label{section7}

This paper proposes a novel, data-driven grasp synthesis approach named \textit{EfficientGrasp}, which generates contact points based on the fingertip workspace of a gripper. Moreover, a model-free RL method is applied to compute the gripper inverse kinematics, further extending the use case of \textit{EfficientGrasp} to novel grippers without requiring a kinematics model. A performance comparison of \textit{EfficientGrasp} and the UniGrasp framework was evaluated in simulation using 3 different grippers. Our proposed method significantly reduces the size of the required gripper data input, decreasing computational memory usage by 81.7\% and producing 9.85\% higher contact points generation accuracy. Our method also shows higher grasping success rate (86.3\%) in simulation when in direct comparison to UniGrasp (83.2\%); and real-world experiments show that UniGrasp
fails when applied to a gripper with closed-loop constraints, while \textit{EfficientGrasp} achieves a success rate of
83.3\%. Experimental results also reveal that the proposed method is able to generate quality grasps in force closure for objects within the physical limits of the applied gripper both in simulation and in the real world. 

In this work, the mass of the objects is assumed to be equally distributed, which implies that a successful grasp can be estimated from object shape. Future work can focus on including a dynamics-based contact points estimator that adapts if an object mass is detected to be unequally distributed after a first grasp attempt.




\bibliographystyle{IEEEtran}
\bibliography{references}

\end{document}